\documentclass{IEEEtran}
\usepackage{amsmath,amssymb,amsfonts}
\usepackage{algorithmic}
\usepackage{graphicx}
\usepackage{textcomp}
\usepackage{wrapfig}
\usepackage[numbers]{natbib}
\usepackage{longtable} 
\usepackage{booktabs} 
\usepackage{tabularx,colortbl}
\usepackage[utf8]{inputenc}
\def\BibTeX{{\rm B\kern-.05em{\sc i\kern-.025em b}\kern-.08em
    T\kern-.1667em\lower.7ex\hbox{E}\kern-.125emX}}
\begin{document}
\title{Hybrid Transformer for Early Alzheimer’s Detection: Integration of Handwriting-Based 2D Images and 1D Signal Features}
\author{Changqing GONG, Huafeng Qin and Mounîm A. El-Yacoubi
\thanks{C. Gong is with Telecom SudParis, Institut Polytechnique de Paris, 91120 Palaiseau, France (e-mail: changqing.gong@telecom-sudparis.eu).}
\thanks{H. Qin is with the School of Computer Science and Information Engineering, Chongqing Technology and Business University, Chongqing 400067, China (e-mail: qinhuafengfeng@163.com).}
\thanks{M. A. El-Yacoubi is Telecom SudParis, Institut Polytechnique de Paris, 91120 Palaiseau, France (e-mail: mounim.el\_yacoubi@telecom-sudparis.eu).}
\thanks{Manuscript received xxx, xx, 2024; revised XXXX XX, 201X. 
This work was supported in part by the xxxx 
(Corresponding author: Huafeng Qin and Mounîm A. El-Yacoubi.)}}
\maketitle
\begin{abstract}
Alzheimer's Disease (AD) is a prevalent neurodegenerative condition where early detection is vital. Handwriting, often affected early in AD, offers a non-invasive and cost-effective way to capture subtle motor changes. State-of-the-art research on handwriting, mostly online, based AD detection has predominantly relied on manually extracted features, fed as input to shallow machine learning models. Some recent works have proposed deep learning (DL)-based models, either 1D-CNN or 2D-CNN architectures, with performance comparing favorably to handcrafted schemes. These approaches, however, overlook the intrinsic relationship between the 2D spatial patterns of handwriting strokes and their 1D dynamic characteristics, thus limiting their capacity to capture the multimodal nature of handwriting data. Moreover, the application of Transformer models remains basically unexplored. To address these limitations, we propose a novel approach for AD detection, consisting of a learnable multimodal hybrid attention model that integrates simultaneously 2D handwriting images with 1D dynamic handwriting signals. Our model leverages a gated mechanism to combine similarity and difference attention, blending the two modalities and learning robust features by incorporating information at different scales. Our model achieved state-of-the-art performance on the DARWIN dataset, with an F1-score of 90.32\% and accuracy of 90.91\% in Task 8 ('L' writing), surpassing the previous best by 4.61\% and 6.06\% respectively. 
\end{abstract}
% Main text
\section{Introduction}\label{Introduction}
Alzheimer’s disease (AD), the most common cause of dementia, is a progressive neurodegenerative disorder (ND) characterized by gradual nerve cell degeneration, leading to cognitive decline in memory, reasoning, and daily functioning \cite{el2019aging(1), singhal2012medicinal(2), alzheimer20192019(3), buchman2011loss(4), armstrong2013criteria(5)}. Similar conditions, including Lewy body disease, frontotemporal degeneration, Parkinson’s disease, and stroke, also impair cognitive functions. The incidence of these diseases increases with age \cite{zhang2021can(6), petersen2018practice[7], ewers2007multicenter(8), baek2022annual(9), nichols2022estimation(10)}. Though incurable, current treatments aim to manage progression, emphasizing the need for improved early diagnostic methods.

The current medical consensus is that dementia is irreversible once clinical symptoms appear, but early detection and intervention can slow its progression \cite{long2019alzheimer(11)}. However, expensive and invasive diagnostics (e.g., A-PET, cerebrospinal fluid testing) \cite{burnham2019application(12)} and subjective neuropsychological tests (e.g., MMSE, MoCA) hinder early diagnosis and widespread screening of Alzheimer’s disease \cite{gosztolya2019identifying(22)}. Researchers have explored biomarkers sensitive to cognitive decline, using machine learning (ML) to analyze signals like eye movement \cite{mengoudi2020augmenting}, speech \cite{mirzaei2018two,garcia2024unveiling}, galvanic skin response \cite{alhassan2017admemento}, and Gait disturbances and frailty \cite{yamada2021combining,hebert2010upper(15), buchman2007frailty(14), boyle2010physical(13)}. Handwriting changes caused by AD have also been studied recently \cite{el2019aging(1), garre2017kinematic(16), kawa2017spatial(17), schroter2003kinematic(18), yan2008alzheimer(21)}. Handwriting, which involves cognitive and motor functions, offers a non-invasive, cost-effective way to track disease progression \cite{yan2008alzheimer(21), impedovo2018dynamic(23), yu2016kinematic(24)}. ML applied to motor function can reduce clinical assessment time \cite{myszczynska2020applications(25)}, and graphic tablets enable easy online handwriting tasks while capturing kinematic and dynamic data \cite{cilia2018experimental(26)}.

State-of-the-art research on handwriting-based AD detection has predominantly relied on manually extracted features, fed as input to shallow ML models\cite{el2019aging(1),kahindo2018characterizing,qi2023study,chai2023classification(34)}. Recently, deep learning (DL) has shown strong feature representation capabilities, yielding promising results in tasks like image segmentation \cite{minaee2021image(27)}, video processing \cite{chen2019distributed(28)}, object tracking \cite{sun2019deep(29)}, and biometric recognition \cite{qin2024attention(30)}. Few works \cite{dao2022detection,mwamsojo2022reservoir,erdogmus2023promise} have proposed, for AD detection, deep learning (DL)-based models, either 1D-CNN modeling 1D feature signals or 2D-CNN modeling 2D handwriting images, outperforming handcrafted schemes. These approaches, however, overlook the relationship between the 2D spatial patterns of handwriting strokes and their 1D dynamic characteristics, thus limiting their capacity to capture the multimodal nature of handwriting data. Moreover, the application of Transformer models remains basically unexplored. Inspired by the success of Transformers in image recognition and natural language processing, we propose a novel hybrid Transformer model for early AD that addresses these limitations. Our Transformer is multimodal as it integrates 2D handwriting images with 1D feature signals, by encoding both modalities and incorporating a learnable similarity and difference attention mechanism. Our model leverages a hybrid attention mechanism and introduces a new loss function combining template contrastive loss with cross-entropy loss to improve classification performance. Designed to be lightweight due to the small dataset size, the model uses a shallower architecture with shorter encodings. We benchmarked our model against state-of-the-art classifiers, achieving superior performance. Our main contributions are as follows:
\begin{itemize}
    \item We propose a novel Transformer-based deep neural network model that enables multi-scale feature representation and outperforms state-of-the-art baselines.
    \item We integrate, within our Transformer model, 1D feature signals with 2D handwriting images. A gating mechanism is employed to blend the similarity and learnable differences between the 2D and 1D features.
    \item We introduce a new loss function, combining template contrastive loss with cross-entropy loss, to learn smoother classification features.
    \item Our model is evaluated on a gold-standard dataset with 25 handwriting tasks, achieving superior performance compared to state-of-the-art classifiers.
\end{itemize}

Next, Section 2 reviews the state of the art. Section 3 outlines the DARWIN dataset tasks and data preprocessing. Section 4 details our proposed model and loss functions. Section 5 presents our experiments, comparing results with state-of-the-art classifiers and analyzing the findings.

\section{Related Work}\label{Related work}

Deterioration in writing ability is a known diagnostic indicator of Alzheimer’s Disease (AD) \cite{garre2017kinematic(31)}, and kinematic handwriting analysis has revealed pathological features in the handwriting process \cite{el2019aging(1)}. Handwriting-based AD detection methods can be broadly categorized into two categories: traditional machine learning (ML) and deep learning (DL).

Many studies have applied traditional ML techniques for AD detection. Qi et al. \cite{qi2023study} used logistic regression on kinematic features, such as writing speed and pen pressure, achieving an accuracy range from 71.5\% to 96.55\%. Chai et al. \cite{chai2023classification(34)} employed SVMs leveraging handwriting dynamics based on writing speed, time, and pressure, with an accuracy of 89\% in distinguishing mild cognitive impairment (MCI) from AD. Meng et al. \cite{meng2022image} applied a 2D discrete Fourier transform, corner detection, and gray-level co-occurrence matrix analysis on Archimedes spiral and labyrinth lattice handwriting images, and achieved a mean AUC of 0.94 with a Decision Tree classifier. Cilia et al. \cite{cilia2022diagnosing} employed Random Forest on a novel large dataset for AD detection, achieving an accuracy of 85.29\%. These methods show promise in identifying temporal dynamics, kinematics, and spatial characteristics associated with Alzheimer's, such as writing speed and letter size. 

Deep learning (DL) has proven to be a powerful tool for handwriting-based neurodegenerative disease detection, including AD and Parkinson’s Disease (PD). DL methods use either 2D image data or 1D feature signals. Given the shortage of papers leveraging DL for assessing AD from handwriting, we report also papers on PD. Pereira et al. \cite{pereira2018handwritten} transformed 1D signals from a smart pen into 2D images for PD classification using CNNs, achieving 93.5\% accuracy. Taleb et al. \cite{taleb2023detection} transformed 1D time series into 2D images, fed to CNN and CNN-BLSTM models, for PD detection. The accuracy improved from 83.33\% to 97.62\% with data augmentation. Diaz et al. \cite{diaz2021sequence} combined 1D convolutional layers with Bi-GRU layers for PD recognition, achieving 94.44\% accuracy.

For AD detection, Cilia et al. \cite{cilia2022diagnosing} introduced the DARWIN (Diagnosis Alzheimer With Handwriting) dataset, with 174 participants, comprising AD patients and healthy controls. In a related study, Cilia et al. \cite{cilia2021online} classified AD using handcrafted and CNN-extracted features from color and binary images. They employed CNN models such as VGG19, ResNet50, InceptionV3, and InceptionResNetV2 to extract features from RGB and binary images, fed to ML algorithms, like k-Nearest Neighbors (kNN), MLP, Random Forest, and SVM, for classification, with CNN-extracted features outperforming handcrafted features. Subsequently, Cilia et al. \cite{cilia2022deep} converted handwriting into color images encoding dynamic information to enhance feature representation. Erdogmus et al. \cite{erdogmus2023promise} transformed manually extracted 1D features into 2D features fed to CNN, achieving an accuracy of 90.4\%. Dao et al. \cite{dao2022detection} developed a 1D-CNN to detect early-stage AD from online handwriting loops. To tackle the limited training data, they employed various data augmentation techniques, including a GAN variant (DoppelGANger) to generate realistic handwriting sequences, achieving an accuracy of 89\% accuracy. It is worth noting that the accuracies reported above are essentially not comparable as most were obtained on different datasets, under different experimental protocols. In our experiments, we implement several state-of-the-art models in order to soundly benchmark our approach on the same dataset.

The literature on handwriting-based AD detection highlights a range of approaches. Traditional ML techniques have been widely adopted by extracting key handwriting features. They often require, however, extensive manual feature engineering, which limits their ability to fully capture the complexity of handwriting variations in AD. DL methods have shown superior performance by learning intricate spatial and dynamic patterns directly from raw handwriting samples. Some studies, nevertheless, still depend on manual feature extraction, converting features into 2D images for CNNs. While a few studies have explored one-dimensional (1D) time series feature signals, none have examined the correlation between 2D handwriting images and 1D signals, and the impact of combining these modalities on AD. Furthermore, the application of Transformers to handwriting recognition for AD remains unexplored, leaving a gap in current research. To address these challenges, we propose a multimodal Transformer model that integrates 2D handwriting images with 1D feature signals, offering a promising approach for more accurate AD detection.
\section{Material and method}\label{Material and method}
In this section, we introduce the dataset, describe our preprocessing of raw signal data, the extraction of 1D signal features and the reconstruction of handwriting images.
\subsection{Dataset}
\begin{figure}[ht]
    \centering
    \includegraphics[width=0.9\columnwidth]{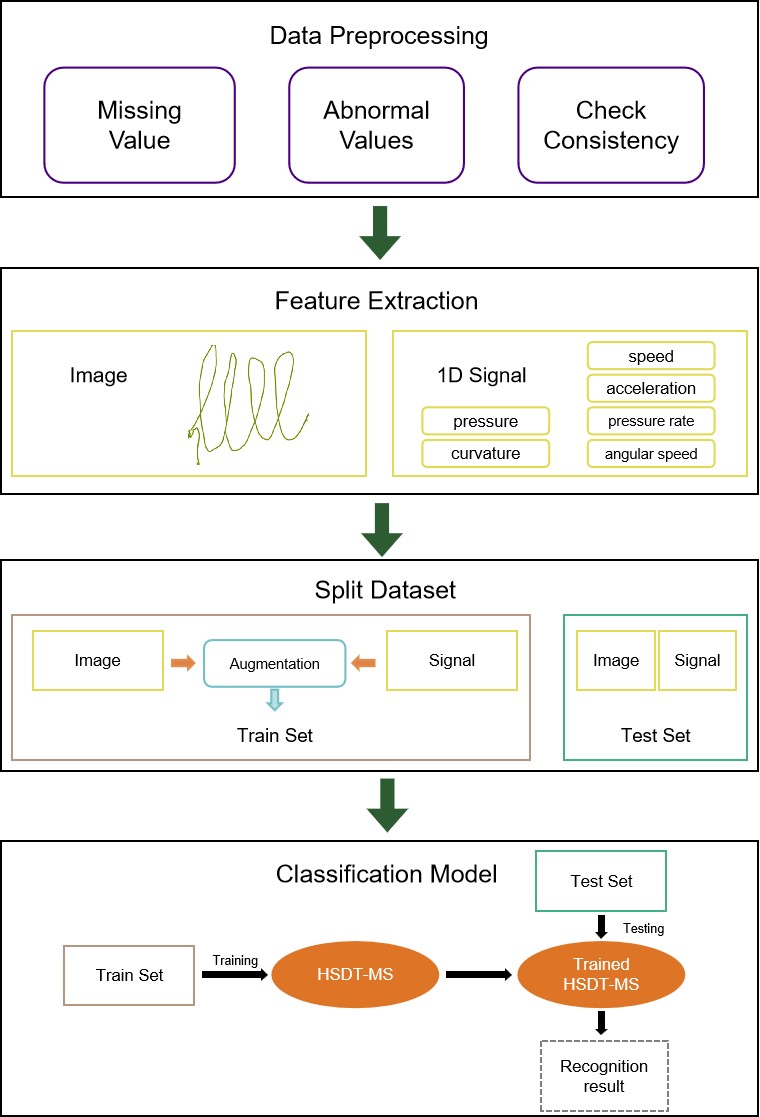}
    \caption{Architecture of HSDT}
    \label{fig:Architectur}
\end{figure}

We used the DARWIN-RAW dataset \cite{cilia2022diagnosing}, a gold-standard resource for AD diagnosis, with data from 174 participants (89 AD patients and 85 healthy controls). This dataset includes 25 handwriting tasks designed for early AD detection \cite{cilia2018experimental}, categorized into four types: (1) graphic tasks, (2) copy tasks, (3) memory tasks, and (4) dictation tasks. The raw handwriting data \((x_i, y_i, p_i)\) were preprocessed to generate 2D images and 1D feature signals. This process was motivated by the effectiveness of kinematic features in detecting early AD. The workflow is illustrated in Figure \ref{fig:Architectur}. 
\subsection{Pre-processing}
In the preprocessing phase, we examined the dataset for missing values based on the timestamps of each handwriting task. Missing values were estimated using interpolation and imputed accordingly, and outliers were removed. If a participant had any unrecorded task, their data for that task were discarded. The data for each task were standardized to have a mean of 0 and a standard deviation of 1.
\subsection{1D Signal Feature Extraction}
\begin{figure}[ht]
    \centering
    \includegraphics[width=0.9\columnwidth]{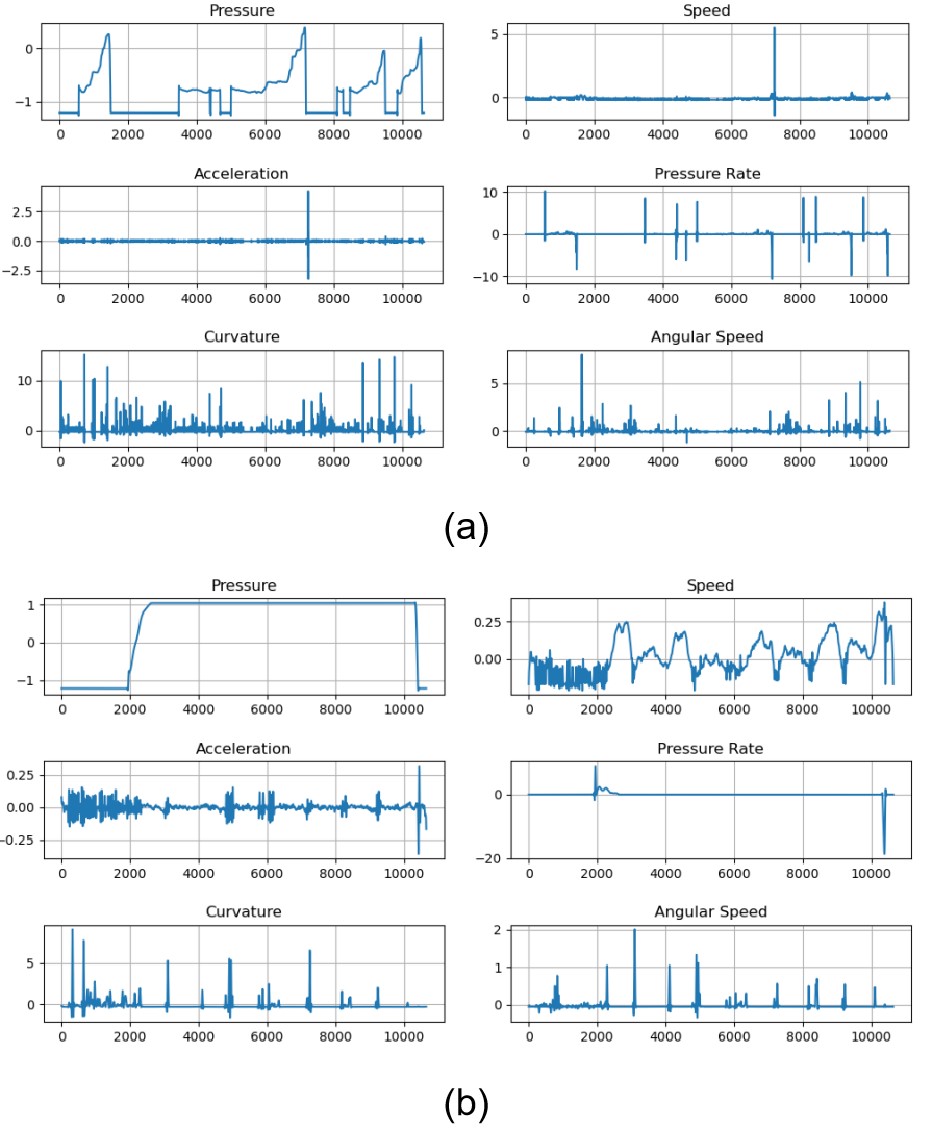}
    \caption{Signal features: (a) Patient's; (b) Healthy Control’s. }
    \label{fig2}
\end{figure}

Six key features, speed, acceleration, pressure rate of change, curvature, and angular speed, were extracted from raw handwriting signals for analysis and model training. The final dataset combines these computed features with the original data. An example of 1D signal is shown in Figure \ref{fig2}.
\subsection{Generation of 2D Images from online handwriting}
\begin{figure}[ht]
    \centering
    \includegraphics[width=0.9\columnwidth]{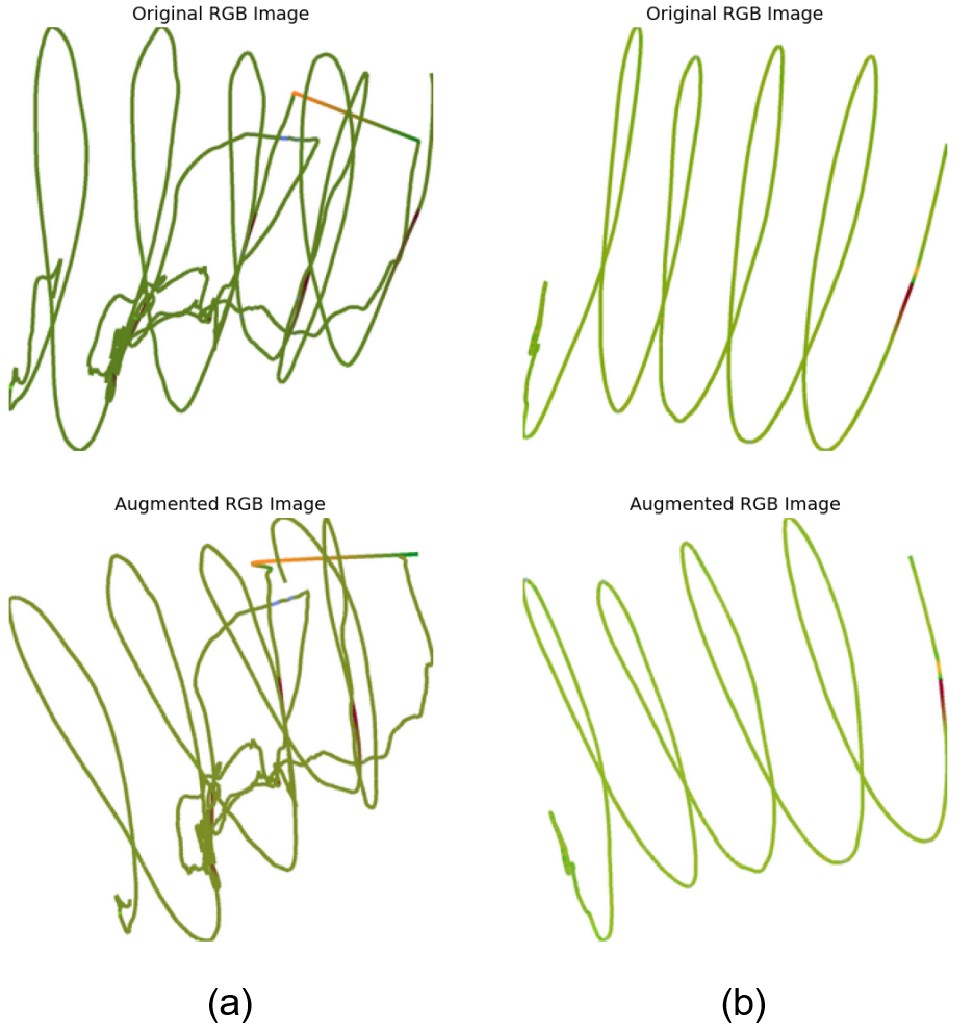}
    \caption{Handwriting images: (a) Patient's; (b) Healthy Control’s.}
    \label{fig3}
\end{figure}

Building on prior work \cite{cilia2022deep}, we used the original \((x_i, y_i)\) coordinates to generate images for network training. In contrast to related studies, the RGB components, \(r_i\), \(g_i\), and \(b_i\), were derived from pressure rate of change, acceleration, and angular velocity, respectively. The generated images were normalized and smoothed using interpolation, with examples shown in Figure \ref{fig3}.
\subsection{Data Augmentation}
To enhance data diversity and improve model generalization, we have applied data augmentation techniques, including rotation, noise addition, scaling, window warping, and window slicing, to simulate real-world disturbances. Some augmented images are shown in Figure \ref{fig3}.

\section{Proposed work}\label{Proposed work}

Handwriting recognition faces challenges in capturing the relationships between handwriting images and signals such as pen pressure, acceleration, and angular velocity, due to the differences between sequence-based 1D signal tasks and vision-based 2D image tasks. To address these challenges, we propose HSDA-MS Transformer, a Multi-Scale Transformer based on Hybrid Similarity and Difference Attention for early Alzheimer's detection. The Hybrid Similarity and Difference Attention (HSDA) scheme employs a gating mechanism to combine similarity and difference weights, capturing dependencies between both 2D images and 1D signals. Convolutions are integrated into the Transformer to capture features at different scales, enhancing robustness. Additionally, we introduce a plug-and-play template contrastive loss function, which updates positive and negative templates during training to learn more discriminant features.

% Traditional convolutions are effective for image tasks by leveraging local receptive fields, while sequence-based data is better suited for LSTM \cite{hochreiter1997long} and RNN \cite{rumelhart1986learning}, which capture temporal dependencies.Transformers have achieved significant success in NLP \cite{vaswani2017attention} and image classification with ViT \cite{dosovitskiy2020image}, emphasizing their potential in visual tasks. The cross-attention mechanism is particularly useful for integrating features from different modalities and scales, enhancing learning. Recent studies, such as DCA \cite{ates2023dual} for medical segmentation, BEFUnet \cite{manzari2024befunet}, and CKD-TransBTS \cite{lin2023ckd}, highlight the effectiveness of cross-attention in medical applications by capturing long-range dependencies and multimodal features. Multimodal image fusion further improves task accuracy by integrating complementary data sources. Examples include TSJNet \cite{jie2024tsjnet}, CoCoNet \cite{liu2024coconet}, and a focus information integration framework \cite{li2024bridging}, which enhance image details and semantic information by fusing multimodal data.

\begin{figure*}[t]
\centerline{\includegraphics[scale=0.60]{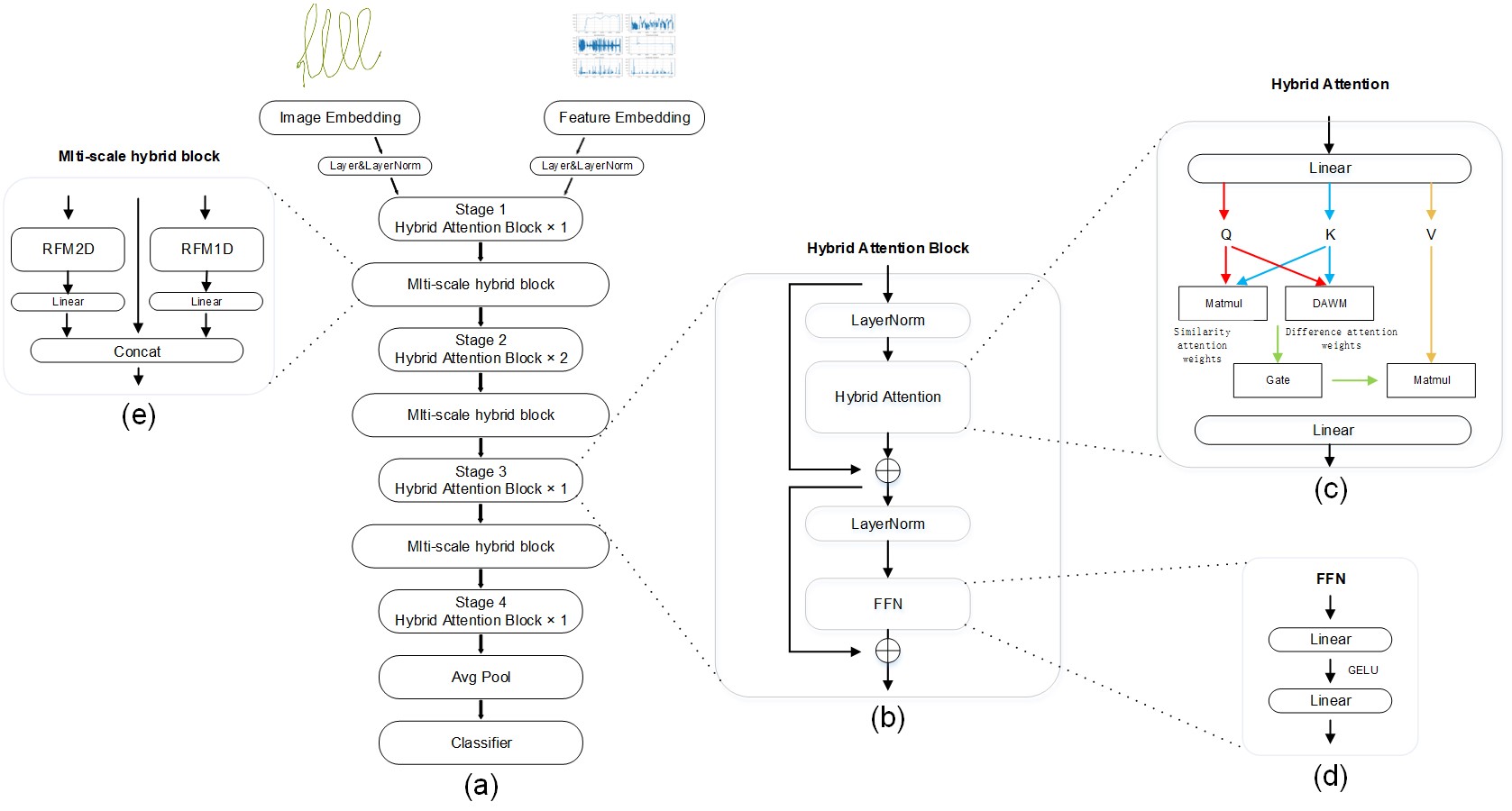}}
\caption{Framework of the HSDA-MS Transformer. }
\label{fig2:Framwork}
\end{figure*}
\subsection{HSDA-MS Transformer}
\subsubsection{Image embedding}
In ViT \cite{dosovitskiy2020image}, an image is split into non-overlapping 2D patches, transformed into 1D embeddings using a multi-layer perceptron (MLP). To preserve spatial information lost in this process, we introduce a stem module. As shown in Figure \ref{fig3:Framwork}(a), the stem block consists of three convolutional layers and an MLP. Three $3\times3$ convolutions with a stride of 2 reduce the input size, while one $3\times3$ convolution with a stride of 1 extracts local spatial features. The MLP then converts the feature map into a fixed-size embedding, capturing global context and abstract features. Given an input image $X_{{2d}} \in {R}^{H\times W \times 3}$, the stem block generates a feature map $X_{{2d}}^{'} \in {R}^{\frac{H}{8}\times\frac{W}{8}\times C}$, where $C=128$. The MLP then transforms it into $X' \in {R}^{1 \times d}$, where $d = 128$, as shown in Eq. \eqref{Stem} and Eq. \eqref{StemMLP}.
\begin{equation}
X_{{2d}}^{'} = {ImageEmbedding} (X_{{2d}})
\label{Stem}
\end{equation}
\begin{equation}
X' = {MLP} (X_{{2d}}^{'})
\label{StemMLP}
\end{equation}
\subsubsection{Signal embedding}
\begin{figure}[!tbp]
\centering
\begin{minipage}[b]{0.4\linewidth}\centering
{\includegraphics[scale=0.8]
{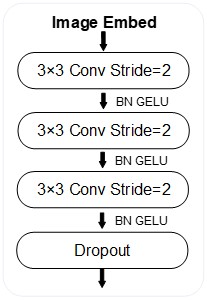}}\\ (a)
\end{minipage}
\begin{minipage}[b]{0.4\linewidth}\centering
{\includegraphics[scale=0.8]
{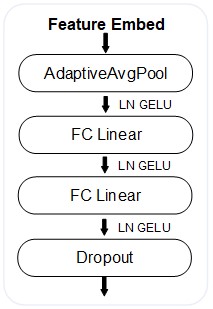}}\\ (b)
\end{minipage}
 \caption{ (a) Image embedding and (b) Signal embedding} 
 \label{fig3:Framwork}
 \end{figure}

In this network, the 1D feature signal is processed to extract robust features. To prevent loss of critical information, we introduce an embedding module using adaptive average pooling, fully connected layers, and normalization. As shown in Figure \ref{fig3:Framwork}(a), the embedding block consists of an adaptive average pooling layer, two fully connected layers, and an MLP. The pooling layer reduces the input signal's dimensionality, while the fully connected layers, with normalization and activation, extract meaningful features. The final MLP converts these features into a fixed-size embedding, capturing global context. Given an input signal $X_{{1d}} \in {R}^{N \times D}$, where $N$ is the number of signals and $D$ is the dimensionality, the embedding block produces $X_{{1d}}^{'} \in {R}^{N \times D' }$ with $D'=2048$. The MLP then transforms this into $X'' \in {R}^{N \times d}$, where $d = 128$, as shown in Eq. \eqref{SignalEmbedding} and Eq. \eqref{SignalEmbeddingMLP}.
\begin{equation}
X_{{1d}}^{'} = {SignalEmbedding} (X_{{1d}})
\label{SignalEmbedding}
\end{equation}
\begin{equation}
X'' = {MLP} (X_{{1d}}^{'})
\label{SignalEmbeddingMLP}
\end{equation}
 \subsubsection{Hybrid Attention block}
The hybrid attention module combines similarity and difference attention, as shown in Figure \ref{fig2:Framwork}(c), with features normalized using a normalization layer \cite{ba2016layer}, followed by a sequential Feed-Forward Network (FFN) to enhance representation, as shown in Figure \ref{fig2:Framwork}(d). The  Multi-scale hybrid module mixes cross-level learning relationships. The 2D features, obtained from upper-layer image features, are processed through three layers of 2D convolution and downsampling to capture multi-scale information. Similarly, 1D features are processed via three layers of 1D convolution and downsampling to obtain high-dimensional signal features. Both 2D and 1D features are concatenated with the output of the hybrid attention module to produce the final feature representation. 

This design offers two advantages: it combines feature differences and similarities for multi-level multimodal feature extraction, and integrates cross-level convolutions to capture both structural and spatial information. This mitigates Transformer's limitations in capturing local relationships and patch-level structural information, promoting comprehensive feature representation learning. Next, we detail the gating mechanism to combine similarity attention and difference attention.

\textbf{Hybrid Attention Module:} As shown in Figure \ref{fig2:Framwork}(c), the proposed hybrid attention model integrates two types of attention: similarity and difference attention, enhancing thereby multimodal feature representation. Similarity attention captures global patterns by focusing on the similarity between queries and keys, providing contextual information. Difference attention, by contrast, learns subtle variations between queries and keys, focusing on local changes. By combining the two, the model captures both global similarities and local differences, allowing for more precise attention distribution. To grant multimodal hybrid attention, feature maps $X'$ and $X''$ are considered as non-overlapping patches and concatenated into $\Bar{X}$. Each patch is transformed into an embedded feature vector, as shown in Eq. \eqref{eq5:concat}:
\begin{equation}
\begin{gathered}
  \Bar{X}=Concat( X', X'')
  \end{gathered}
 \label{eq5:concat}
\end{equation}
the feature map $\Bar{X}$ is converted into a token sequence $\Bar{X} \in {R}^{\Bar{N} \times d}$, where $\Bar{N} = N + 1$ represents the number of patches. Subsequently, $\Bar{X}$ is transformed through three linear layers, resulting in three matrices: $Q$, $K$, and $V$. The matrices \(Q\), \(K\), and \(V\) are the query, key, and value matrices, calculated as \(Q = \Bar{X}W_Q\), \(K = \Bar{X}W_K\), and \(V = \Bar{X}W_V\). 

\textbf{Similarity attention weights}: To perform similarity attention among $\Bar{N}$ tokens, we use the dot product between the $Q$ and $K$ tokens to calculate the similarity attention weights as follows Eq. \eqref{eq6:SAW}:
\begin{equation}
SAW(Q, K) = {Softmax}\left(\frac{Q \cdot K^{T}}{\sqrt{d}} + B\right)
\label{eq6:SAW}
\end{equation}
where $B\in R^{\Bar{N} \times \Bar{N}}$ indicates the relative position bias, $Softmax(\cdot)$ is applied to the rows of the similarity matrix $A=QK^T$ with $d$ providing normalization. 

\textbf{Difference attention weights}: Inspired by graph convolutional networks\cite{kipf2016semi, velivckovic2017graph, xu2018powerful}, where relationships between nodes are learned by calculating differences between input nodes, we propose, in this work, a feature difference attention mechanism that captures local differences and provides fine-grained feature information. This allows the model to more accurately adjust attention distribution and identify subtle changes in input data. 
\begin{figure}[ht]
\centerline{\includegraphics[scale=0.8]{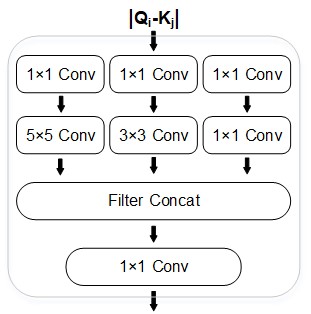}}
\caption{Discrepancy Attention Weight Model. }
\label{fig4:Framwork}
\end{figure}
As shown in Figure \ref{fig4:Framwork}, after computing the absolute differences between the \( Q \) and \( K \) matrices, we employ convolutional blocks with different kernel sizes to aggregate the discrepancy weights among adjacent nodes. This aims to capture the diverse discrepancy information between different nodes. A MLP is then used to update the discrepancy information and learn the correlations between them. For each element in query matrix \( Q \), the absolute difference with every element in key matrix \( K \) is computed. These differences are then fed into the Discrepancy Attention Weight Model to obtain the discrepancy attention weights (DAW) between the query and the key, as shown in Eq. ~\eqref{eq7:matrix} and Eq. ~\eqref{eq7:DAW}:
\begin{equation}
D_{i, j} = \left| Q_i - K_j \right|
\label{eq7:matrix}
\end{equation}
\begin{equation}
DAW(Q, K) = {Softmax}\left( M_{\theta}\left( \left| Q_i - K_j \right| \right)\right)
\label{eq7:DAW}
\end{equation}
where \( | Q_i - K_j | \) denotes the absolute difference between the \( i \)-th element of matrix \( Q \) and the \( j \)-th element of matrix \( K \), \( D_{i, j} \) is a discrepancy matrix, \( \{ D_{i, j} \in {R}^{\Bar{N} \times \Bar{N} \times d} \mid i, j = 1, \ldots, \Bar{N} \} \), and \( M_{\theta} \) represents the aggregation of information using convolutional blocks with kernel sizes of 5, 3, and 1, as shown in Figure \ref{fig5}. These convolutional blocks capture information from the neighboring nodes. 
\begin{figure}[ht]
\centerline{\includegraphics[scale=0.4]{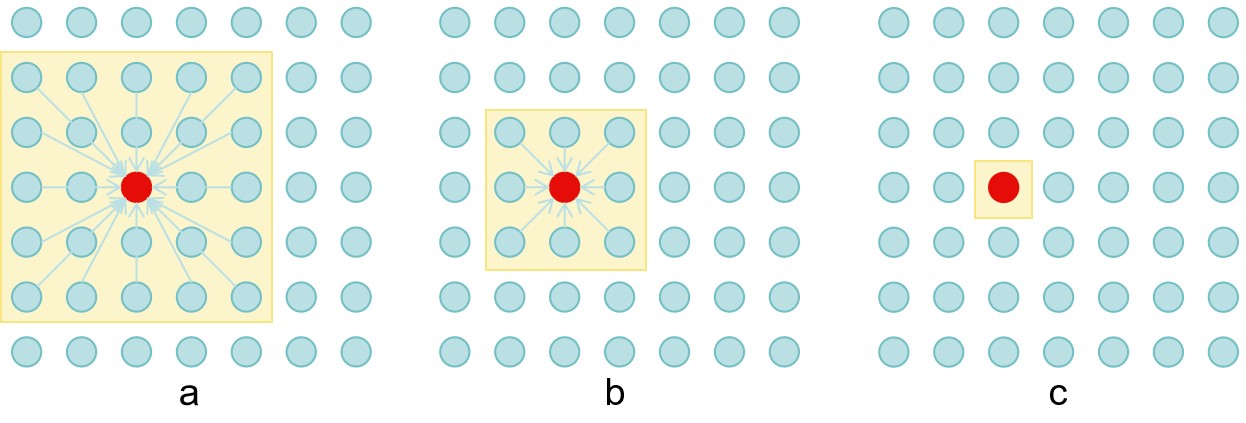}}
\caption{Aggregate DI (discrepancy information): a) 2-neighbor DI, b) 1-neighbor DI, c) self-DI. }
\label{fig5}
\end{figure}

\textbf{Gating Mix:} To aggregate the value matrix \( V \) using the attention weights for the updated feature representation, we combine the similarity attention weights and the discrepancy attention weights, as shown in Eq. ~\eqref{eq8:HA}:
\begin{equation}
\begin{gathered}
HA=Mix(SAW(Q, K), DAW(Q, K))V
 \end{gathered}
 \label{eq8:HA}
\end{equation}
where $Mix(\cdot)$ is a gating mixing operation. The gating mechanism learns gating weights, allowing the model to flexibly adjust the proportion of the two attention weights based on different input features. This dynamic adjustment helps capture the diversity and complexity of the input data. The $Mix(\cdot)$ function, based on the inputs \(Q\) and \(K\), is formulated by Eq. \eqref{eq9:mix} and Eq. \eqref{eq10:G}:
\begin{equation}
{Mix}(SAW, DAW) = G \cdot SAW + (1 - G) \cdot DAW
 \label{eq9:mix}
\end{equation}
\begin{equation}
G = \sigma(W_g [SAW; DAW])
 \label{eq10:G}
\end{equation}
where \(\sigma\) is the Sigmoid function, \(W_g\) is a learnable weight matrix, \([SAW; DAW]\) denotes the concatenation of \(SAW\) and \(DAW\), respectively. To capture enriched information, we concatenate the $L$ individual attention heads to construct a multi-head attention, as shown in Eq. \eqref{eq11:msat}:
\begin{equation}
\begin{gathered}
  \Bar{X}'=Concat( HA_{1}, HA_2, . . . , HA_L)W
  \end{gathered}
 \label{eq11:msat}
\end{equation}
where $HA_{h} = Mix(SAW_{h}, DAW_{h})V_{h}$, and $h$ indicates the head number. 

To facilitate description, we pack all equations in the mix
attention process into Eq. \eqref{eq12}:
\begin{equation}
\begin{gathered}
\Bar{X}'=HSDT(\Bar{X})
 \end{gathered}
\label{eq12}
\end{equation}

\textbf{FFN:} The FFN is a two-layer feed-forward neural network applying non-linear transformations to enhance feature extraction, as shown in Eq. \eqref{eq14}:
\begin{equation}
\Bar{X}'' = MLP(MLP(\Bar{X}'))
\label{eq14}
\end{equation}

Based on Eq. (\ref{eq12}) and Eq. (\ref{eq14}), (as shown in Figure \ref{fig2:Framwork}(b)), we restate them as Eq. \eqref{eq16} and Eq. \eqref{eq17} respectively:
\begin{equation}  
\begin{gathered}
\bar{Y^{l}_{i}}=\bar{X^{l}_{i}}+HSDT(LN(\bar{X^{l}_{i}})), 
\end{gathered}
\label{eq16}
\end{equation}
\begin{equation}  
\begin{gathered}
\bar{X}^{l}_{i+1}=\bar{Y^{l}_{i}}+FFN(LN(\bar{Y^{l}_{i}})). 
 \end{gathered}
\label{eq17}
\end{equation}
where $l$ represents the number of stages, as shown in Figure \ref{fig2:Framwork}(a), with $l \in (1, 2, 3, 4)$, and $i$ denotes the number of blocks.

\subsubsection{Mlti-scale hybrid block}
Multi-scale feature fusion leverages information from different scales to extract richer features. Methods such as Feature Pyramid Networks (FPN) \cite{lin2017feature}, BiFPN \cite{tan2020efficientdet}, YoloV3 \cite{redmon2018yolov3}, Inception \cite{szegedy2015going}, and PSPnet \cite{zhao2017pyramid} achieve feature fusion by introducing hierarchical structures and fusion techniques. Transformer-based models, such as HVT \cite{pan2021scalable}, PVT \cite{wang2021pyramid}, and MViT \cite{fan2021multiscale}, have incorporated pyramid structures into ViT to improve performance. Recently, Qin et al. \cite{qin2023label} proposed a Multi-Scale Vein Transformer (MSVT) to learn dependencies between patches at different scales, while also integrating convolutions to enhance robustness.

Handwriting patterns in Alzheimer's patients often exhibit irregularities and fine-grained tremors, as shown in Figure \ref{fig2}. To model these tremors, we propose a multi-scale module that extracts features at different scales for each layer of the hybrid attention module. Capturing detailed features at various scales enhances the model's robustness and generalization.

\textbf{2D Residual Feedforward Module:} RFM2D is a residual block \cite{he2016deep} where the traditional convolution learns a feature representation over a localized receptive field by the convolution kernels, with weights shared over the whole feature map. The intrinsic characteristics of a locality mechanism allows information exchange within a local region. Specifically, we first pool the feature map $X_{{2d}}^{l}$ obtained from the previous layer. Within the first multi-scale feature map fusion, $X_{{2d}}^{l}$ refers to the feature map obtained from Eq. \eqref{Stem}. Fine-grained features are then extracted as shown in Eq. \eqref{eq18:Down} and Eq. \eqref{eq19:rfm2d}:
\begin{equation}
\begin{gathered}
Y^{l}_{2d}=P(X_{{2d}}^{l})
 \end{gathered}
 \label{eq18:Down}
\end{equation}
\begin{equation}
\begin{gathered}
X^{l+1}_{2d}=Y^{l}_{2d}+Conv_{1\times 1}(DWConv(Conv(Y^{l}_{2d})))
 \end{gathered}
 \label{eq19:rfm2d}
\end{equation}
where $P(\cdot)$ is a 2D convolution with a kernel size of 3 and a stride of 2. The separable convolution DWConv(·) extracts local information with minimal additional computational cost. Similar to classical residual networks, the residual connection enhances the gradient propagation capability across layers. Then, we flatten the feature map $X^{l+1}_{2d}$ obtained from Eq. \eqref{eq19:rfm2d}, and use an MLP to extract high-dimensional features, as shown in Eq. \eqref{eq20:amp2d}:
\begin{equation}
\begin{gathered}
Z'=MLP(X^{l+1}_{2d})
 \end{gathered}
 \label{eq20:amp2d}
\end{equation}

\textbf{1D Residual Feedforward Module:} We first pool the feature map $X_{{1d}}^{l}$ obtained from the previous layer. During the first multi-scale feature map fusion, $X_{{1d}}^{l}$ refers to the feature map obtained from Eq. \eqref{SignalEmbedding} and then extract fine-grained features as shown in Eq. \eqref{eq21:Down} and Eq. \eqref{eq22:rfm1d}:
\begin{equation}
\begin{gathered}
Y^{l}_{1d}=P(X_{{1d}}^{l})
 \end{gathered}
 \label{eq21:Down}
\end{equation}
\begin{equation}
\begin{gathered}
X^{l+1}_{1d}=Y^{l}_{1d}+Conv1d(DWConv1d(Conv1d(Y^{l}_{1d})))
 \end{gathered}
 \label{eq22:rfm1d}
\end{equation}
where $P(\cdot)$ is an adaptive 1D max pooling layer. Then, we flatten the feature map $X^{l+1}_{1d}$ obtained from Eq. \eqref{eq22:rfm1d}, and use an MLP to extract high-dimensional features (Eq. \eqref{eq23:amp1d}). 
\begin{equation}
\begin{gathered}
Z''=MLP(X^{l+1}_{1d})
 \end{gathered}
 \label{eq23:amp1d}
\end{equation}

Based on Equations \eqref{eq20:amp2d} and \eqref{eq23:amp1d}, we concatenate \(Z'\) and \(Z''\), and then concatenate the result with the output of the hybrid attention module. This concatenated result is used as input for the next stage of the hybrid attention module. To facilitate the description, we refer to the output of the hybrid attention module as \(Z'''\), 
 as shown in Eq. \eqref{eq24:concat}:
\begin{equation}
\begin{gathered}
  \Bar{Z}=Concat( Z',Z'',Z''')
  \end{gathered}
 \label{eq24:concat}
\end{equation}
where $\Bar{Z} \in {R}^{\Bar{N} \times (d + d')}$, $\Bar{N} = N + 1,$ and $ d' $ is the vector length of $Z'$ and $Z''$. 

% \begin{equation}
% \Bar{Z}'= Concat( \Bar{Z}, HA)
% \label{eq25:concat}
% \end{equation}

% The Multi-scale Hybrid Block generates a feature map $\Bar{Z}' \in {R}^{\Bar{N} \times (d + d')}$. 

\subsection{Template contrastive loss}
The motivation for introducing the template contrastive loss is to enhance the model's ability to distinguish between positive and negative samples by explicitly learning from their differences. By incorporating both cross-entropy and contrastive losses, we aim to leverage the benefits of supervised classification while ensuring that the model learns robust, discriminative features. This approach helps in creating a clearer separation in the feature space, leading to improved classification performance and better generalization to unseen data. The adaptive update mechanism for template vectors further refines the model's learning process, making it more responsive to the nuances of the data distribution. 

The template vectors for positive and negative samples, denoted as \(\mathbf{T}_p\) and \(\mathbf{T}_n\), are initialized with a dimensionality \(d\) by sampling from a standard normal distribution \(\mathcal{N}(0, I)\), where \(d\) corresponds to the dimensionality of the feature vectors in the layer preceding the final classification layer.

\textbf{Cross-entropy loss:} The cross-entropy loss, used for supervised classification tasks, is defined as Eq. \eqref{eq:24}:
\begin{equation}
\mathcal{L}_{{CE}} = - \frac{1}{N} \sum_{i=1}^{N} \sum_{c=1}^{C} y_{i, c} \log \hat{y}_{i, c}
\label{eq:24}
\end{equation}
where \(N\) is the batch size, \(C\) is the number of classes, \(y_{i, c}\) is the true label of sample \(i\), and \(\hat{y}_{i, c}\) is the predicted probability distribution by the model. 

\textbf{Contrastive Loss:} Given feature vector \(\mathbf{f}_i \in {R}^d\) and label \(y_i\) for sample \(i\), template vectors \(\mathbf{T}_p\) and \(\mathbf{T}_n\), and the number of samples \(N\), the Contrastive Loss is defined as Eq. \eqref{eq25}:
% \begin{equation}
% \begin{gathered}
% \mathcal{L}_{\text{contrastive}} = \frac{1}{N} \sum_{i=1}^{N} \left( y_i \cdot (d_p^i + (\text{m} - d_n^i)) + (1 - y_i) \cdot ( d_n^i + (\text{m} - d_p^i)) \right)
% \end{gathered}
% \end{equation}
\begin{equation}
\begin{gathered}
\mathcal{L}_{{contrastive}} = \frac{1}{N} \sum_{i=1}^{N} \left( y_i \cdot d_p^i + (1 - y_i) \cdot d_n^i  \right)
\label{eq25}
\end{gathered}
\end{equation}
where  $d_p^i = 1 -  {cosine\_similarity}(\mathbf{f}_i, \mathbf{T}_p)$, $d_n^i = 1 -  {cosine\_similarity}(\mathbf{f}_i, \mathbf{T}_n)$, \(y_i\) is the label of sample \(i\) (positive sample is 1, negative sample is 0). This formula integrates the calculation methods for both positive and negative sample contrastive losses. The total loss, combining the above two losses, is defined as Eq. \eqref{eq26}:
\begin{equation}
\mathcal{L}_{{total}} = \mathcal{L}_{{CE}} + \lambda \cdot \mathcal{L}_{{contrastive}}
\label{eq26}
\end{equation}
where\(\lambda\) is an adjustable hyperparameter used to control the relative weight of the cross-entropy loss and contrastive loss in the total loss. In our experiments, \(\lambda\) is set to 0. 8. 

\textbf{Template updated:} At the end of each batch, the template vectors are updated based on both the feature vectors of the current batch and the templates from the previous batch:
\begin{equation}
\mathbf{T}_p^{(k+1)} = \alpha \mathbf{T}_p^{(k)} + (1 - \alpha) \frac{1}{|P|} \sum_{i \in P} \mathbf{f}_i
\end{equation}
\begin{equation}
\mathbf{T}_n^{(k+1)} = \alpha \mathbf{T}_n^{(k)} + (1 - \alpha) \frac{1}{|N|} \sum_{i \in N} \mathbf{f}_i
\end{equation}
where \(\alpha\) is a smoothing factor, \(\alpha\) is set to 0.9, and \(|P|\) and \(|N|\) are the number of positive and negative samples in the current batch, respectively. 
\section{Experimental results and discussion}
In this section, we present the experimental setup, performance evaluation metrics, recognition performance results, and ablation studies.
\subsection{Experimental Setup}
To assess our approach, we conducted extensive experiments on the DARWIN-RAW publicly available gold-standard dataset, collected using Wacom's Bamboo tablet from 174 participants. The x-y coordinate sequences of pen-tip movements were recorded at a frequency of 200 Hz. The dataset consists of x-y coordinates (174 subjects \( \times \) 25 tasks \( \times \) 1 x-y coordinate sequence, with some missing data). The x-y coordinates were then processed and augmented following the procedures described in Chapter 3. We compared our model's classification performance against various state-of-the-art classifiers, including CNN-2D(AD)\cite{erdogmus2023promise}, CNN-1D(AD)\cite{dao2022detection}, VGG\cite{simonyan2014very}, ResNet\cite{he2016deep}, 
DenseNet\cite{huang2017densely}, Inception-ResNetV2\cite{szegedy2017inception}, 
Xception\cite{chollet2017xception}, and MobileNetV2\cite{sandler2018mobilenetv2}. For a fair comparison, we used pretrained models from the TIMM library. During training, we set the learning rate to 0.01 and the batch size to 16. The optimizer used was Stochastic Gradient Descent (SGD) with a momentum parameter of 0. 9 and a weight decay parameter of 0.05. Additionally, we employed cosine annealing as the learning rate scheduler and set the maximum number of training epochs to 100, with early stopping, halting the training when the accuracy did not improve for 10 consecutive epochs. All experiments were conducted using the PyTorch framework on a computer equipped with NVIDIA\texttrademark GPUs. 

\subsection{Evaluation Metrics}
We employed standard evaluation metrics, namely Accuracy, Precision, Recall (also known as Sensitivity), and F1-score, to assess our model classification performance. Let \( P \) denote the positive samples, the samples labeled with the target class (AD), and \( N \) denote the negative samples, labeled as HC. 
Accuracy is the most widely-used evaluation metric, representing the ratio of correctly predicted samples to the total number of samples. Precision is the ratio of correctly predicted positive samples to the total samples predicted as positive. Recall is the ratio of correctly predicted positive samples to all actual positive samples. The F1-score, the harmonic mean of Precision and Recall, is particularly useful for evaluating performance on imbalanced datasets. 

% The equations for these metrics are provided in Eqs. \eqref{eq32:accuracy}-\eqref{eq35:f1-score}. 
% \begin{equation}
% \text{Accuracy} = \frac{TP + TN}{TP + TN + FP + FN}
% \label{eq32:accuracy}
% \end{equation}
% \begin{equation}
% \text{Precision} = \frac{TP}{TP + FP}
% \label{eq33:precision}
% \end{equation}
% \begin{equation}
% \text{Recall} = \frac{TP}{TP + FN}
% \label{eq34:recall}
% \end{equation}
% \begin{equation}
% \text{F1-score} = 2 \cdot \frac{{Precision} \cdot {Recall}}{{Precision} + {Recall}}
% \label{eq35:f1-score}
% \end{equation}

\subsection{Recognition Performance for HSDT}
\begin{table}%[]
\caption{Performance Comparison on Task 1, Task 2, and Task 5}\label{tbl1}
\begin{tabularx}{\columnwidth}{@{}lXXXXX@{}}
\toprule
Model & F1score & Accuracy & Precision & Recall  \\ 
\midrule
\multicolumn{5}{c}{\textbf{Task 1}} \\
\textbf{Ours}      & \textbf{81.08} & \textbf{79.41} & \textbf{75.00} & \textbf{88.24} \\ 
VGG19              & 60.61  & 61.77  & 62.50   & 58.82  \\ 
ResNet152          & 76.47  & 76.47  & 76.47   & 76.47  \\ 
DenseNet201        & 78.95  & 76.47  & 71.43   & 88.24  \\ 
InceptionResNetV2  & 80.00  & 76.47  & 69.57   & 94.12  \\ 
Xception41         & 70.59  & 70.59  & 70.59   & 70.59  \\ 
MobileNetV2        & 72.22  & 70.59  & 68.42   & 76.47  \\ 
CNN-2D(AD)            & 62.50  & 64.71  & 66.67   & 58.82  \\ 
CNN-1D(AD)            & 68.42  & 64.71  & 61.91   & 76.47  \\ 
VIT                & 48.00  & 61.77  & 75.00   & 35.29  \\ 
PVT                & 64.71  & 64.71  & 64.71   & 64.71  \\ 
SwinTransformer    & 73.17  & 67.65  & 62.50   & 88.24  \\ 
\midrule
\multicolumn{5}{c}{\textbf{Task 2}} \\
\textbf{Ours}      & \textbf{83.87} & \textbf{84.85} & \textbf{92.86} & \textbf{76.47} \\ 
VGG19              & 59.26  & 66.67  & 80.00   & 47.06  \\ 
ResNet152          & 76.47  & 75.76  & 76.47   & 76.47  \\ 
DenseNet201        & 80.00  & 81.82  & 92.31   & 70.59  \\ 
InceptionResNetV2  & 78.05  & 72.73  & 66.67   & 94.12  \\ 
Xception41         & 70.97  & 72.73  & 78.57   & 64.71  \\ 
MobileNetV2        & 70.97  & 72.73  & 78.57   & 64.71  \\ 
CNN-2D(AD)            & 68.97  & 72.73  & 83.33   & 58.82  \\ 
CNN-1D(AD)            & 78.79  & 78.79  & 81.25   & 76.47  \\ 
VIT                & 66.67  & 72.73  & 90.00   & 52.94  \\ 
PVT                & 73.33  & 75.76  & 84.62   & 64.71  \\ 
SwinTransformer    & 81.25  & 81.82  & 86.67   & 76.47  \\ 
\midrule
\multicolumn{5}{c}{\textbf{Task 5}} \\
\textbf{Ours}      & \textbf{87.50} & \textbf{87.88} & \textbf{93.33} & \textbf{82.35} \\ 
VGG19              & 78.79  & 78.79  & 81.25   & 76.47  \\ 
ResNet152          & 84.85  & 84.85  & 87.50   & 82.35  \\ 
DenseNet201        & 74.29  & 72.73  & 72.22   & 76.47  \\ 
InceptionResNetV2  & 73.33  & 75.76  & 84.62   & 64.71  \\ 
Xception41         & 70.97  & 72.73  & 78.57   & 64.71  \\ 
MobileNetV2        & 72.73  & 63.64  & 59.26   & 94.12  \\ 
CNN-2D(AD)            & 76.47  & 75.76  & 76.47   & 76.47  \\
CNN-1D(AD)            & 70.27  & 66.67  & 65.00   & 76.47  \\ 
VIT                & 70.97  & 72.73  & 78.57   & 64.71  \\ 
PVT                & 80.00  & 81.82  & 92.31   & 70.59  \\ 
SwinTransformer    & 76.47  & 75.76  & 76.47   & 76.47  \\ 
\bottomrule
\end{tabularx}
\end{table}

\begin{table}%[]
\caption{Performance Comparison on Task 8, Task 17, and Task 24}\label{tbl2}
\begin{tabularx}{\columnwidth}{@{}lXXXXX@{}}
\toprule
Model              & F1score & Accuracy & Precision & Recall  \\ 
\midrule
\multicolumn{5}{c}{\textbf{Task 8}} \\
\textbf{Ours}      & \textbf{90.32} & \textbf{90.91} & \textbf{100.00} & \textbf{82.35} \\ 
VGG19              & 85.71  & 84.85  & 83.33   & 88.24  \\ 
ResNet152          & 82.76  & 84.85  & 100.00  & 70.59  \\ 
DenseNet201        & 80.00  & 81.82  & 92.31   & 70.59  \\ 
InceptionResNetV2  & 81.25  & 81.82  & 86.67   & 76.47  \\ 
Xception41         & 78.57  & 81.82  & 100.00  & 64.71  \\ 
MobileNetV2        & 75.86  & 78.79  & 91.67   & 64.71  \\ 
CNN-2D(AD)            & 77.42  & 78.79  & 85.71   & 70.59  \\ 
CNN-1D(AD)            & 75.00  & 75.76  & 80.00   & 70.59  \\
VIT                & 72.73  & 72.73  & 75.00   & 70.59  \\ 
PVT                & 80.00  & 81.82  & 92.31   & 70.59  \\ 
SwinTransformer    & 75.86  & 78.79  & 91.67   & 64.71  \\ 
\midrule
\multicolumn{5}{c}{\textbf{Task 17}} \\
\textbf{Ours}      & \textbf{86.49} & \textbf{84.85} & \textbf{80.00} & \textbf{94.12} \\ 
VGG19              & 75.68  & 72.73  & 70.00   & 82.35  \\ 
ResNet152          & 81.08  & 78.79  & 75.00   & 88.24  \\ 
DenseNet201        & 80.95  & 75.76  & 68.00   & 100.00 \\ 
InceptionResNetV2  & 74.29  & 72.73  & 72.22   & 76.47  \\ 
Xception41         & 80.00  & 78.79  & 77.78   & 82.35  \\ 
MobileNetV2        & 78.79  & 78.79  & 81.25   & 76.47  \\ 
CNN-2D(AD)            & 78.05  & 72.73  & 66.67   & 94.12  \\
CNN-1D(AD)            & 72.73  & 72.73  & 75.00   & 70.59  \\
VIT                & 80.00  & 81.82  & 92.31   & 70.59  \\ 
PVT                & 77.78  & 75.76  & 73.68   & 82.35  \\ 
SwinTransformer    & 70.97  & 72.73  & 78.57   & 64.71  \\ 
\midrule
\multicolumn{5}{c}{\textbf{Task 24}} \\
\textbf{Ours}      & \textbf{76.92} & \textbf{81.25} & \textbf{100.00} & \textbf{62.50} \\  
VGG19              & 64.00  & 71.88  & 88.89   & 50.00  \\ 
ResNet152          & 76.47  & 75.76  & 76.47   & 76.47  \\ 
DenseNet201        & 71.43  & 75.00  & 83.33   & 62.50  \\ 
InceptionResNetV2  & 68.97  & 71.88  & 76.92   & 62.50 \\ 
Xception41         & 75.00  & 75.00  & 75.00   & 75.00  \\ 
MobileNetV2        & 75.00  & 75.00  & 75.00   & 75.00  \\ 
CNN-2D(AD)            & 70.59  & 68.75  & 66.67   & 75.00  \\ 
CNN-1D(AD)            & 73.33  & 75.00  & 78.57   & 68.75  \\
VIT                & 64.29  & 68.75  & 75.00   & 56.25  \\ 
PVT                & 58.33  & 68.75  & 87.50   & 43.75  \\ 
SwinTransformer    & 69.23  & 75.00  & 90.00   & 56.25  \\ 
\bottomrule
\end{tabularx}
\end{table}

We evaluated the performance of existing methods across six subtask datasets, encompassing four task categories: memory and dictation (M), graphic (G), and copy (C). Due to the similarity among several tasks within the 25 subtasks, we selected a representative subset of these subtasks. As described in Section 5, 20\% of the entire dataset was set aside as the test set. The remaining data were used for training and validation purposes, according to the stratified k-fold cross-validation technique, that maintains the percentage of samples for each class. Based on the experimental results of the hyperparameter optimization, $k$ was set to 4, meaning the training set was divided into 4 parts: the first part used as the validation set, and the remaining 3 parts used as the training set. This process was repeated 4 times, utilizing the entire dataset for both training and validation. Table\ref{tbl1} and Table\ref{tbl2} present the recognition performance of the various methods on each subtask dataset. 

\begin{figure}[htbp]
    \centering
    \includegraphics[width=\columnwidth]{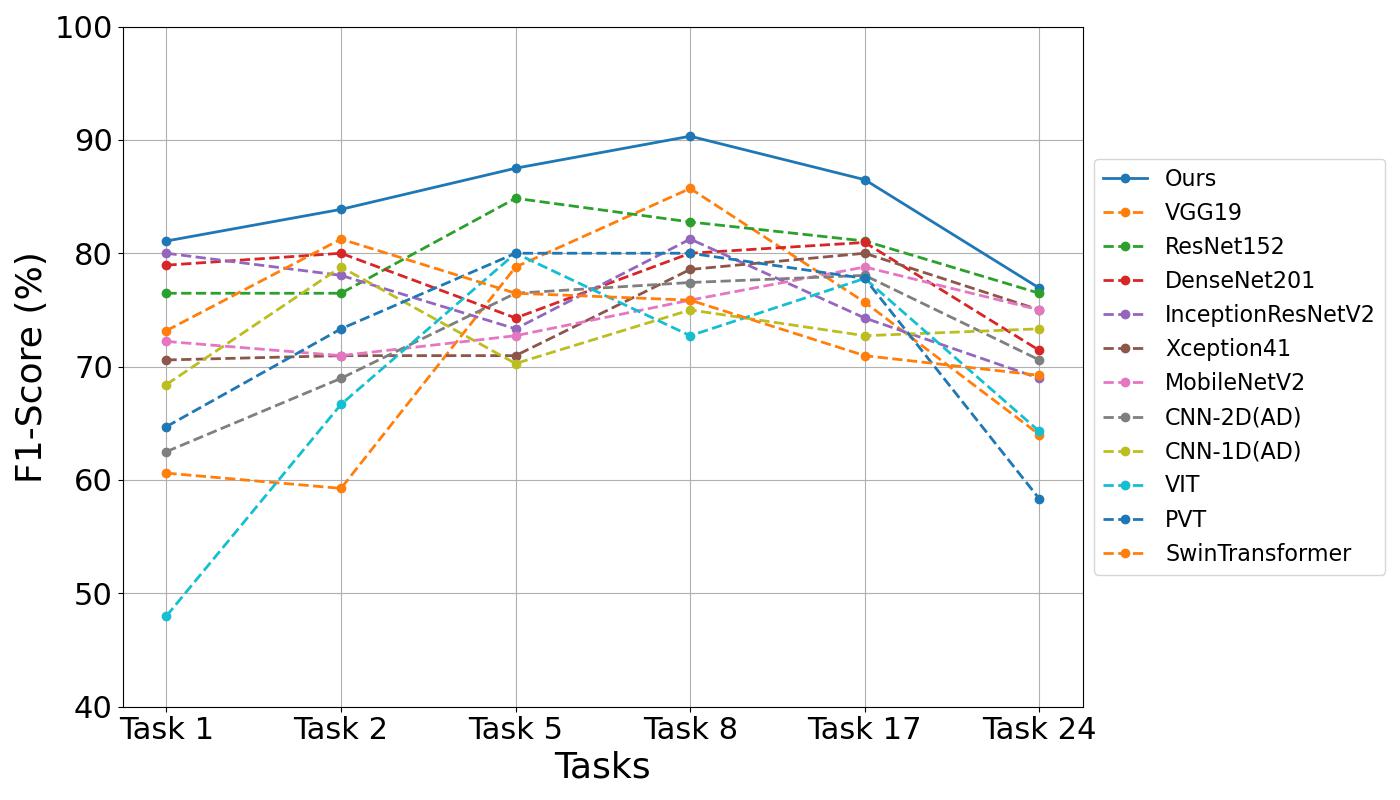}
    \caption{F1-Score Comparison of 12 Models Across 6 Tasks. }
    \label{fig8}
\end{figure}

\begin{figure}[htbp]
    \centering
    \includegraphics[width=\columnwidth]{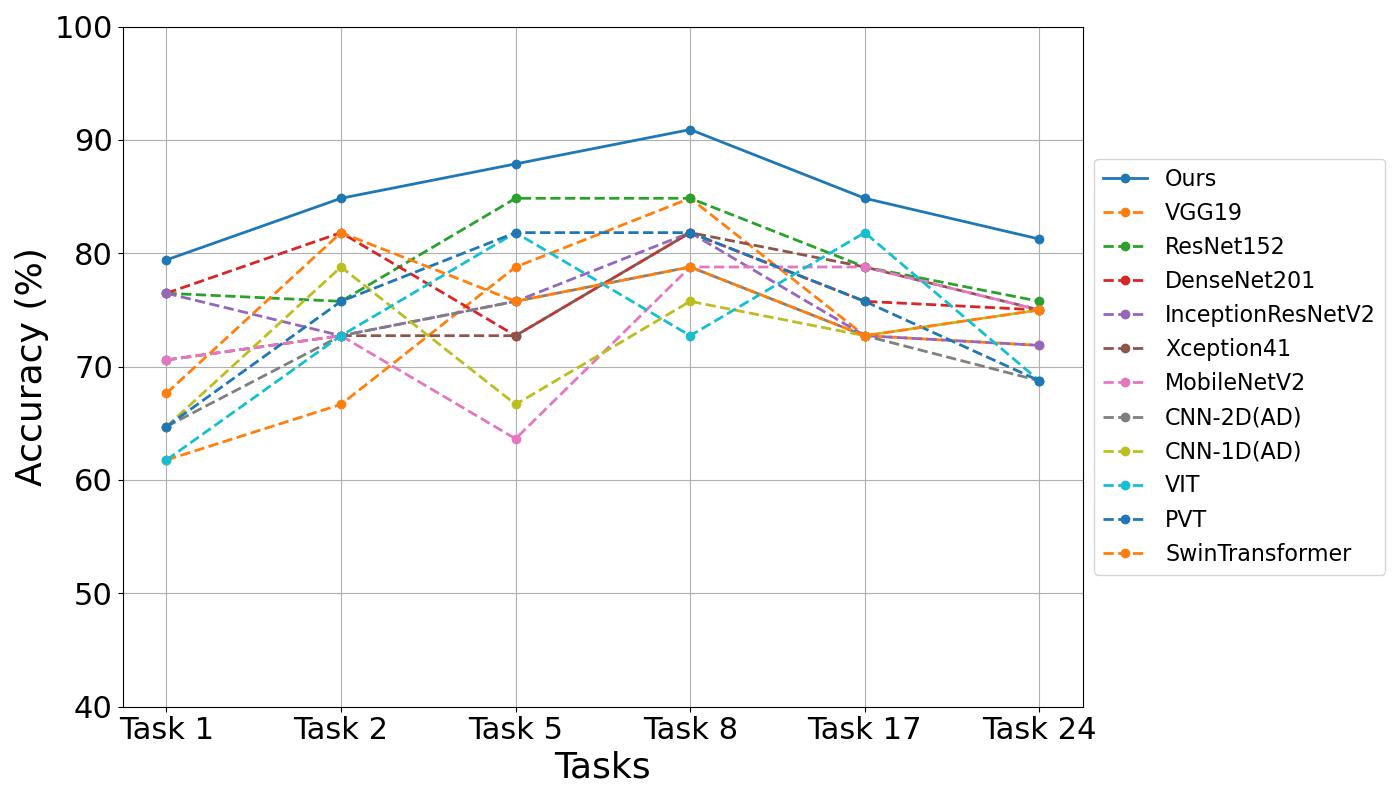}
    \caption{Accuracy Comparison of 12 Models Across 6 Tasks. }
    \label{fig9}
\end{figure}

The results shown in Table \ref{tbl1} and Table \ref{tbl2} clearly demonstrate that our method significantly outperforms existing classifiers across multiple tasks. Specifically, our approach achieved the highest recognition accuracy on sub-datasets Task 1, Task 2, Task 5, Task 8, Task 17, and Task 24, with accuracies of 79.41\%, 84.85\%, 87.88\%, 90.91\%, 84.85\%, and 78.13\%, respectively. These results underscore the robustness and effectiveness of our model in achieving superior classification performance across varied datasets. The results are visualized in Figure \ref{fig8} and Figure \ref{fig9}.
%The confusion matrices of our model on each task are shown in Figure \ref{fig10}.

Furthermore, the F1-scores reflect the balance between precision and recall, both of which are critical in evaluating classification models, particularly in imbalanced datasets. Our method consistently demonstrated high F1-scores across several tasks, achieving 87.50\%, 90.32\%, and 86.49\% on Task 5, Task 8, and Task 17, respectively. These high F1-scores indicate that our approach not only excels in accuracy but also ensures a balanced trade-off between the correct identification of positive instances and minimizing false positives. The performance across these tasks highlights the model's ability to generalize well while maintaining reliability across different data distributions. 


% /%\begin{figure*}[htbp]
% /%    \centering
%     \includegraphics[width=\textwidth]{figs/output.png}
%     \caption{Confusion Matrices. }
%     \label{fig10}
% \end{figure*}

In addition to these achievements, our approach demonstrated superior performance in terms of recall, particularly on Task 1 and Task 8, where it identified a significant proportion of positive instances without compromising precision. This further affirms the model's applicability in real-world scenarios where both false negatives and false positives have significant impacts. 

This superior performance can be attributed to Four key factors: 1) Combining Feature Similarity and Difference: the hybrid attention module integrates similarity attention and difference attention, leveraging the strengths of both. Similarity attention captures global patterns by computing the dot product of queries and keys, providing global context information that aids in recognizing global patterns and dependencies within the input sequences. Difference attention, on the other hand, focuses on the differences between queries and keys, capturing local feature variations, particularly excelling at detecting local features and changes. By combining these two attention mechanisms, the model not only captures global patterns and dependencies but also finely adjusts the attention distribution to recognize subtle variations in the input data, thereby enhancing the model's adaptability to complex data and tasks; 2) Dynamic Adjustment Mechanism: the Mix function within the hybrid attention module utilizes a gating mechanism that flexibly adjusts the proportion of similarity attention and difference attention based on the input features. This dynamic adjustment mechanism helps the model better capture the diversity and complexity of the input data, thereby enhancing the model's adaptability; 3) Cross-Level Feature Learning Capability: the hybrid scale module processes 2D and 1D features independently through convolution and downsampling, generating high-dimensional feature maps of different sizes. These features are then concatenated with the output features of the hybrid attention module. This cross-level feature learning approach effectively integrates relationships learned from different levels, enabling the model to better capture multi-level features in multimodal input data. This design not only enriches the expressiveness of feature representations but also allows for an effective capture of multimodal features. This cross-level convolution processing compensates for the Transformer model's limitations in handling local relationships and block-level structural information, facilitating interaction between features of different scales and contributing to comprehensive feature representation learning; 4) Integration of Local and Global Information: by integrating local and global information, the model more effectively utilizes the structural and spatial information present in the input data. Difference attention further processes local differences through convolution modules, capturing differential information between neighboring nodes, while similarity attention provides global context. This combination enhances the model's comprehensiveness and accuracy when handling multimodal inputs. 
\subsection{Ablation Study Results}
\begin{table}%[]
\caption{Ablation Study: Effect of Multi-scale Hybrid Block and Template Contrastive Loss}\label{tbl_ablation}
\begin{tabularx}{\columnwidth}{@{}lXXXXX@{}}
\toprule
Model & F1score & Accuracy & Precision & Recall  \\ 
\midrule
\multicolumn{5}{c}{\textbf{Task 1}} \\
HSDT                   & \textbf{81.08} & \textbf{79.41} & \textbf{75.00} & \textbf{88.24} \\ 
HSDT without MSH        & 76.47  & 76.47  & 76.47   & 76.47  \\ 
HSDT without CL         & 75.68  & 73.53  & 70.00   & 82.35  \\ 
\midrule
\multicolumn{5}{c}{\textbf{Task 2}} \\
HSDT                   & \textbf{83.87} & \textbf{84.85} & \textbf{92.86} & \textbf{76.47} \\ 
HSDT without MSH        & 78.79  & 78.79  & 81.25   & 76.47  \\ 
HSDT without CL         & 77.42  & 78.79  & 85.71   & 70.59  \\ 
\midrule
\multicolumn{5}{c}{\textbf{Task 5}} \\
HSDT                   & \textbf{87.50} & \textbf{87.88} & \textbf{93.33} & \textbf{82.35} \\ 
HSDT without MSH        & 76.47  & 76.47  & 76.47   & 76.47  \\ 
HSDT without CL         & 78.95  & 75.76  & 71.43   & 88.24  \\ 
\midrule
\multicolumn{5}{c}{\textbf{Task 8}} \\
HSDT                   & \textbf{90.32} & \textbf{90.91} & \textbf{100.00} & \textbf{82.35} \\ 
HSDT without MSH        & 85.71  & 84.85  & 83.33   & 88.24  \\ 
HSDT without CL         & 85.71  & 84.85  & 83.33   & 88.24  \\ 
\midrule
\multicolumn{5}{c}{\textbf{Task 17}} \\
HSDT                   & \textbf{86.49} & \textbf{84.85} & \textbf{80.00} & \textbf{94.12} \\ 
HSDT without MSH        & 80.00  & 81.82  & 92.31   & 70.59  \\ 
HSDT without CL         & 85.71  & 84.85  & 83.33   & 88.24  \\ 
\midrule
\multicolumn{5}{c}{\textbf{Task 24}} \\
HSDT                   & \textbf{76.92} & \textbf{81.25} & \textbf{100.00} & \textbf{62.50} \\ 
HSDT without MSH        & 74.07  & 78.13  & 90.91   & 62.50  \\ 
HSDT without CL         & 73.33  & 75.00  & 78.57   & 68.75  \\ 
\bottomrule
\end{tabularx}
\end{table}

\begin{figure*}[htbp]
    \centering
    \includegraphics[width=\textwidth]{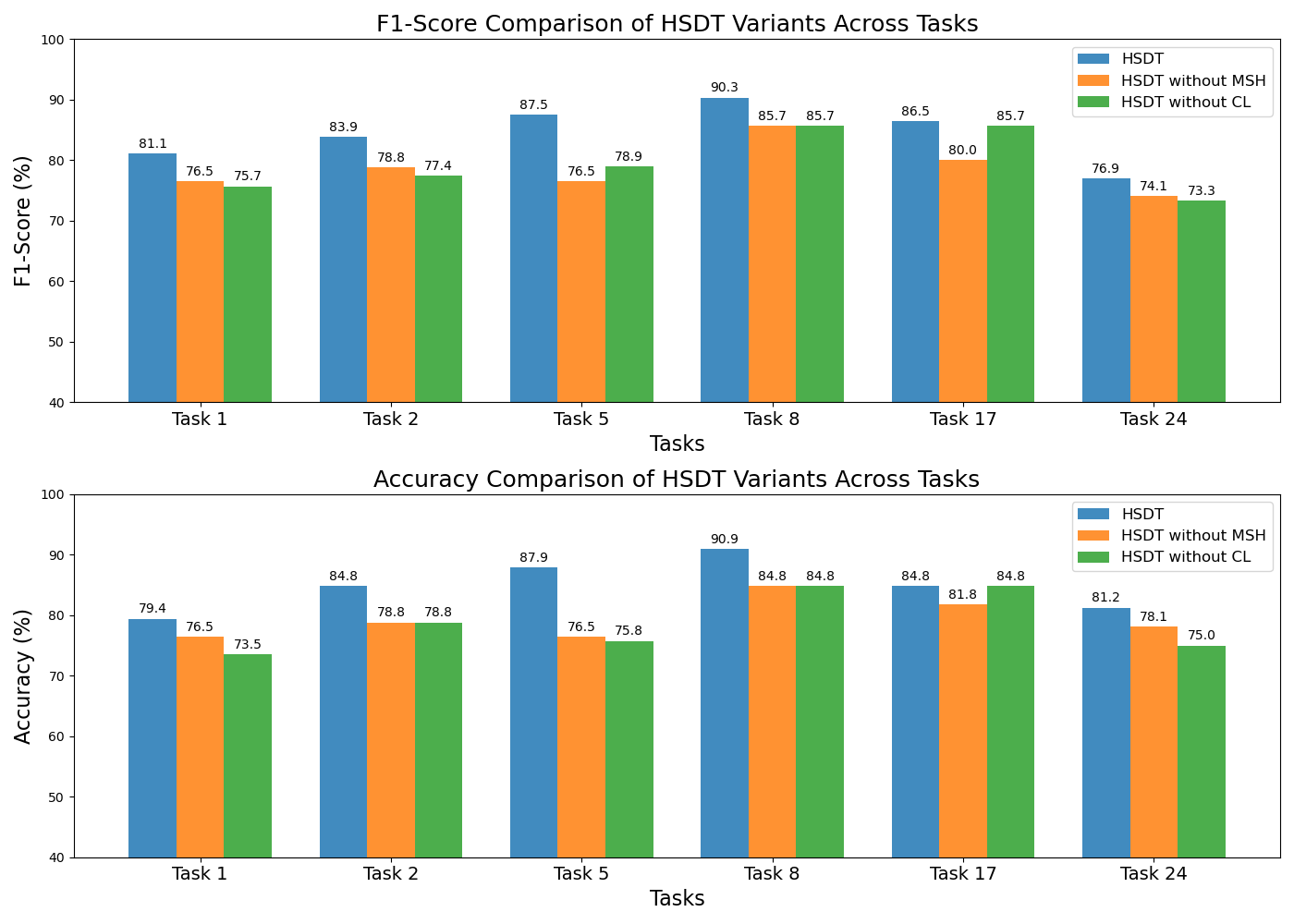}
    \caption{Ablation Study: F1-Score and Accuracy Comparison Across Tasks. }
    \label{fig11}
\end{figure*}

We conducted two ablation studies on the six sub-datasets, one without the Multi-scale Hybrid Block (MSH), and the other without using the Template Contrastive Loss (CL). The experimental results are shown in Table \ref{tbl_ablation}. 

The results demonstrate the significant impact of both components on model’s performance. Specifically, the removal of the Multi-scale Hybrid Block led to a notable decline in both accuracy and F1-score across all tasks. For instance, on Task 1, the F1-score dropped from 81.08\% to 76.47\%, and similar trends were observed in other tasks. This confirms that the Multi-scale Hybrid Block plays a crucial role in capturing multi-level features, which are essential for distinguishing subtle patterns in the data. The visualization of the ablation study is shown in Figure \ref{fig11}. 

The Multi-scale Hybrid Block integrates multi-scale features from both 1D signal data and 2D images, enabling the model to capture local fine-grained details as well as global contextual patterns. This is particularly important for tasks involving complex data, such as handwriting signals, where both small variations in stroke patterns and overarching movement trends need to be considered. The block's ability to fuse features across different scales allows the model to better generalize across tasks, contributing to the model’s robustness and enhanced classification performance. 

Moreover, by utilizing cross-level feature fusion, the Multi-scale Hybrid Block enables the model to learn more comprehensive representations, which enhances its ability to detect subtle distinctions between healthy controls and patients with AD. Without this component, the model’s capacity to process both local and global information simultaneously is weakened, leading to lower accuracy and F1-scores, as observed in the ablation results. 

In the second ablation study, the removal of the Template Contrastive Loss also resulted in a significant decrease in performance, particularly in precision and F1-score. For example, in Task 5, the precision dropped from 93.33\% to 71.43\%, and the F1-score decreased from 87.50\% to 78. 95\%. This highlights the critical role of the Template Contrastive Loss in enhancing feature discrimination. 

The Template Contrastive Loss boosts the model’s ability to learn more robust and discriminative representations by explicitly modeling the similarity relationships between samples in high-dimensional space. By enforcing a separation between positive and negative samples, it ensures that the learned features are more distinct, leading to better classification outcomes. This is particularly important in datasets with overlapping or ambiguous class boundaries, where the contrastive loss helps the model to better differentiate between the subtle patterns associated with AD and normal aging. 

Additionally, the dynamic template update mechanism within the Template Contrastive Loss allows the model to continuously refine its understanding of the feature space throughout the training process, improving adaptability and generalization. The ablation study clearly demonstrates that removing this component diminishes the model’s ability to accurately classify challenging cases, as evidenced by the drop in precision and overall performance. 

In conclusion, the ablation study results underscore the importance of both the Multi-scale Hybrid Block and Template Contrastive Loss. Together, these components enhance the model’s ability to capture complex, multi-scale features and improve feature discrimination, leading to more accurate and robust classification across a range of tasks. 

% \begin{thebibliography}{9}
% \bibitem{fpn}
% T. Lin, P. Dollár, R. Girshick, K. He, B. Hariharan, S. Belongie, "Feature Pyramid Networks for Object Detection, " \textit{Proceedings of the IEEE Conference on Computer Vision and Pattern Recognition (CVPR)}, 2017. 
% \end{thebibliography}

% % Figure
% \begin{figure}%[]
%   \centering
% %    \includegraphics{}
%     \caption{}\label{fig1}
% \end{figure}

% \begin{table}%[]
% \begin{tabular*}{\tblwidth}{@{}LL@{}}
% \toprule
% \midrule
%  & \\
%  & \\
%  & \\
%  & \\
% \bottomrule
% \end{tabular*}
% \end{table}

% Uncomment and use as the case may be
% \begin{theorem} 
% \end{theorem}

% Uncomment and use as the case may be
% \begin{lemma} 
% \end{lemma}

% \appendix
\section{Conclusion}
In this study, we propose a novel HSDA-MS Transformer model for early detection of Alzheimer’s Disease (AD). The model integrates both 2D handwriting images and 1D dynamic signal data, effectively capturing global and local feature variations. It demonstrates strong performance across multiple handwriting tasks by introducing a hybrid similarity and difference attention mechanism, a multi-scale hybrid block, and a template contrastive loss function, all validated through rigorous data processing and experimental evaluation.

The hybrid similarity and difference attention mechanism allows the model to capture both global patterns, such as stroke structure, and subtle local variations, crucial for detecting AD-related motor impairments. The similarity attention mechanism focuses on global handwriting patterns, while the difference attention mechanism refines the detection of fine-grained changes, improving the model's ability to process complex multimodal data.

The multi-scale hybrid block further enhances feature representation by incorporating information from multiple scales. By fusing features from different levels of both 2D and 1D modalities, the model captures fine local details and broad global patterns, resulting in improved classification performance. This multi-scale approach strengthens the model’s ability to handle the complexities of handwriting tasks and adapt to varied input conditions.

The template contrastive loss function enhances the model’s ability to discriminate between AD patients and healthy controls. By comparing positive and negative samples and learning their relationships in high-dimensional space, the loss function improves class separation, leading to more accurate classifications and better generalization to new data. This ensures the model can effectively distinguish early-stage AD from normal aging patterns.

In conclusion, the HSDA-MS Transformer model successfully integrates the hybrid similarity and difference attention mechanism, multi-scale hybrid block, and template contrastive loss function to achieve superior performance in early AD detection. Future work could explore applying this model to other neurodegenerative diseases and extending its use within multimodal deep learning frameworks, potentially integrating additional data types, such as EEG or speech analysis, for broader clinical applications. We will also investigate sound explainability techniques to uncover which patterns in the handwriting inputs are most predictive of AD \cite{sweidan2024explainability}

\section{Declaration of competing interest}
The authors declare that they have no known competing financial interests or personal relationships that could have appeared to influence the work reported in this paper.

% \appendix
% \section{My Appendix}
% Appendix sections are coded under \verb+\appendix+. 

% \verb+\printcredits+ command is used after appendix sections to list 
% author credit taxonomy contribution roles tagged using \verb+\credit+ 
% in frontmatter. 


\bibliographystyle{cas-model2-names}

\bibliography{cas-refs}

% \vskip10pt

\begin{IEEEbiography}[{\includegraphics[width=1in,height=1.25in,clip,keepaspectratio]{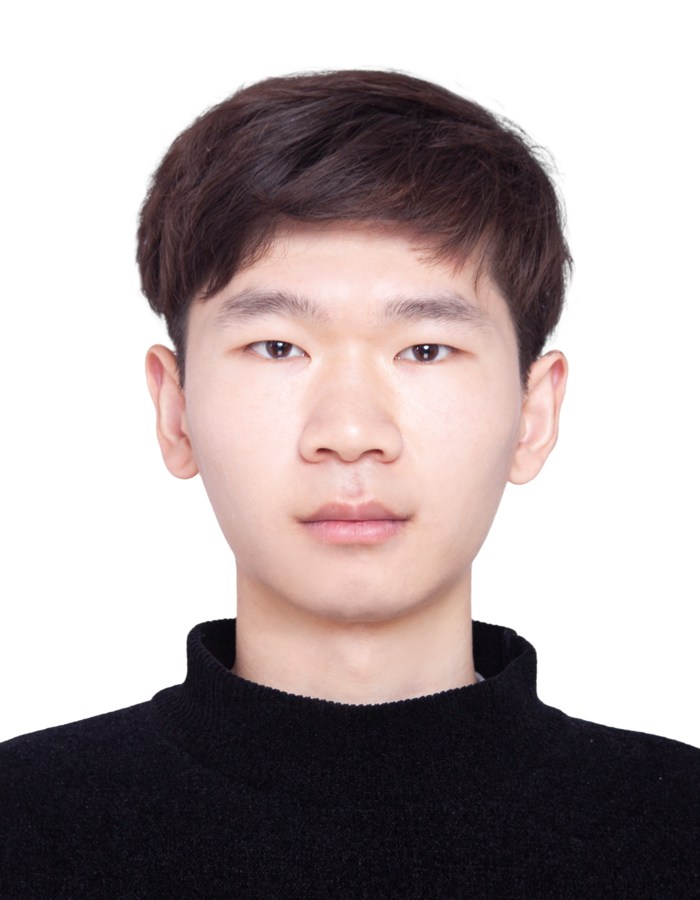}}]{Changqing GONG} 

% \begin{wrapfigure}{l}{0.3\columnwidth} % l 表示图片位于左侧
%     \centering
%     \includegraphics[width=0.3\columnwidth]{figs/changqing.jpg}
% \end{wrapfigure}
received his Bachelor's degree in Software Engineering from Zhongyuan University of Technology in June 2020, and his Master's degree from the Chongqing Key Laboratory of Intelligent Perception and Blockchain Technology at Chongqing Technology and Business University in June 2023. He is currently pursuing a Ph.D. at Institut Polytechnique de Paris. His research interests include vein recognition, biometrics, and machine learning.
\end{IEEEbiography}

\begin{IEEEbiography}[{\includegraphics[width=1in,height=1.25in,clip,keepaspectratio]{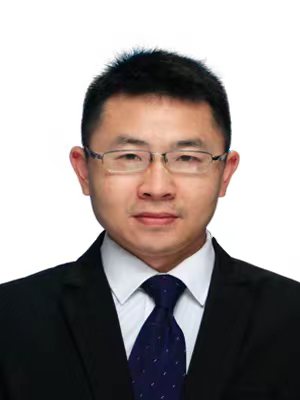}}]{Huafeng Qin} 

% \begin{wrapfigure}{l}{0.3\columnwidth} % l 表示图片位于左侧
%     \centering
%     \includegraphics[width=0.3\columnwidth]{figs/Huafeng_Qin.jpeg}
% \end{wrapfigure}
received BSc degree from the School of Mathematics and Physics and MEng degree from the College of Electronic and Automation from Chongqing University of Technology, and the PhD degree from the College of Opto-Electronic Engineering, Chongqing University. He was a visiting student for 12 months with Nanyang Technological University, Singapore, and then a postdoctoral researcher for two years with Université Paris-saclay, France. Currently, he is a professor with the School of Computer Science and Information Engineering, Chongqing Technology and Business University, China. His research interests include Biometrics (e.g., vein, face, and gait) and machine learning.
\end{IEEEbiography}

% \vspace{10cm}
\begin{IEEEbiography}[{\includegraphics[width=1in,height=1.25in,clip,keepaspectratio]{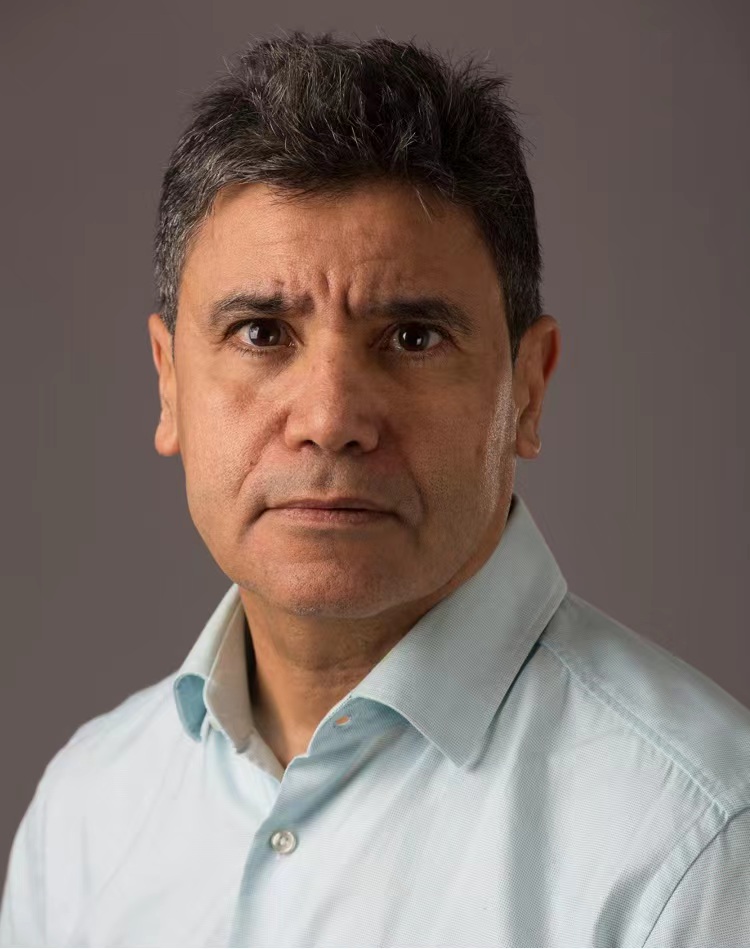}}]{Mounîm A. El-Yacoubi} 

% \begin{wrapfigure}{l}{0.3\columnwidth} % l 表示图片位于左侧
%     \centering
%     \includegraphics[width=0.3\columnwidth]{figs/Mounim_AEl-Yacoubi.jpeg}
% \end{wrapfigure}
received the PhD degree from the University of Rennes, France, in 1996. He was with the Service de Recherche Technique de la Poste (SRTP) with Nantes, France, from 1992 to 1996, where he developed software for Handwritten Address Recognition that is still running in Automatic French mail sorting machines. He was a visiting scientist for 18 months with the Centre for Pattern Recognition and Machine Intelligence (CENPARMI) in Montreal, Canada, and then an associate professor (1998-2000) with the Catholic University of Parana (PUC-PR) in Curitiba, Brazil. From 2001 to 2008, he was a senior software engineer with Parascript, Boulder (Colorado, USA), a world leader company in automatic processing of handwritten and printed documents (mail, checks, forms), for which he developed real-life software for address and check recognition. Since June 2008, he has been a Professor with Telecom SudParis, University of Paris Saclay. His main interests include machine learning, human gesture and activity recognition, human robot interaction, video surveillance and biometrics, information retrieval, and handwriting analysis and recognition.
\end{IEEEbiography}

\end{document}